\documentclass[journal]{IEEEtran}
\usepackage{amsmath,amsfonts}
\usepackage{algorithm}
\usepackage{algpseudocode}
\usepackage{array}
\usepackage{tikz}
\usetikzlibrary{tikzmark}
\usepackage{adjustbox}
\usepackage{booktabs}
\usepackage{textcomp}
\usepackage{caption}
\usepackage{stfloats}
\usepackage{url}
\usepackage{verbatim}
\usepackage{graphicx}
\usepackage{cite}
\usepackage{multirow}
\def\BibTeX{{\rm B\kern-.05em{\sc i\kern-.025em b}\kern-.08em
    T\kern-.1667em\lower.7ex\hbox{E}\kern-.125emX}}
\usepackage{balance}
\usepackage[hidelinks, pagebackref=true, breaklinks=true, colorlinks, bookmarks=false]{hyperref}

\usepackage{soul}
\soulregister\cite7 
\soulregister\citep7 
\soulregister\citet7 
\soulregister\ref7 
\soulregister\pageref7 
\usepackage{bbding}
\usepackage{subcaption}
\usepackage{makecell}
\usepackage{colortbl}
\definecolor{mygray}{gray}{.9}
\definecolor{mypink}{rgb}{.99,.91,.95}
\definecolor{mycyan}{cmyk}{.3,0,0,0}
\usepackage[pagebackref=true,breaklinks=true, colorlinks,bookmarks=false]{hyperref}
\newcommand{\fakeparagraph}[1]{\vspace{1mm}\noindent\textbf{#1}}

\begin{document}

\title{TUMTraf EMOT: Event-Based Multi-Object Tracking Dataset and Baseline for Traffic Scenarios}

\author{Mengyu Li\,$^{2}$, Xingcheng Zhou\,$^{2}$, Guang Chen\,$^{3}$, Alois Knoll\,$^{2}$,~\IEEEmembership{Fellow,~IEEE}, Hu Cao\,$^{1,2*}$,~\IEEEmembership{Member,~IEEE}
\thanks{$*$ Corresponding author.}
\thanks{Authors Affiliation: $^{1}$School of Automation, Southeast University,
$^{2}$Chair of Robotics, Artificial Intelligence and Real-time Systems, Technical University of Munich, Munich, Germany,
		$^{3}$Department of Computer Science and Technology, Tongji University, Shanghai, China.}}

\markboth{Journal of \LaTeX\ Class Files,~Vol.~14, No.~8, August~2021}%
{Shell \MakeLowercase{\textit{et al.}}: A Sample Article Using IEEEtran.cls for IEEE Journals}


\maketitle

\begin{abstract}

In Intelligent Transportation Systems (ITS), multi object tracking is mainly based on frame based cameras. However, these cameras often suffer from performance degradation under dim lighting conditions and fast motion, which are common in real world driving scenarios. Event cameras provide an alternative sensing modality with low latency, high dynamic range and high temporal resolution, making them well suited for safety critical vehicle perception tasks. Despite these advantages, existing event based vision studies in ITS are still limited in scale, and most available datasets focus on detection rather than multi object tracking. To address this gap, we introduce an initial pilot dataset designed for event based ITS applications, covering vehicle and pedestrian detection and tracking in urban traffic scenes. Based on this dataset, we provide a reference tracking by detection benchmark with an improved detector integrated into a standard tracking pipeline, aiming to serve as a baseline for future research on event based multi object tracking. 
\end{abstract}


\section{Introduction}

\IEEEPARstart{F}{or} multi-object tracking (MOT) in Intelligent Transportation Systems (ITS), the primary task is to detect and track multiple objects as they move through a sequence of frames in a video~\cite{Zhao2022MovingOD}. As the most critical participants in ITS, vehicles and pedestrians require precise representation. Traditional frame-based cameras often fall short in environments with extreme exposure or dim lighting conditions, where important visual details can be completely lost. They also struggle in capturing high-speed movements, as their frame capture mechanism combined with limited dynamic range often results in motion blur~\cite{Chen2020EventBasedNV}. In contrast, neuromorphic cameras (event cameras) have the ability to compensate for the above-mentioned shortcomings of RGB cameras. Instead of capturing entire frames at fixed intervals by conventional cameras, event cameras independently detect brightness variations at each pixel. This capability allows them to follow moving objects with less blur. Moreover, because event cameras respond to changes in log-intensity rather than absolute brightness, making the video quality inherently robust to different lighting conditions~\cite{Chen2020EventBasedNV,ESurvey}.

\begin{figure}[!t]
    \centering
    \includegraphics[width=3.5 in]{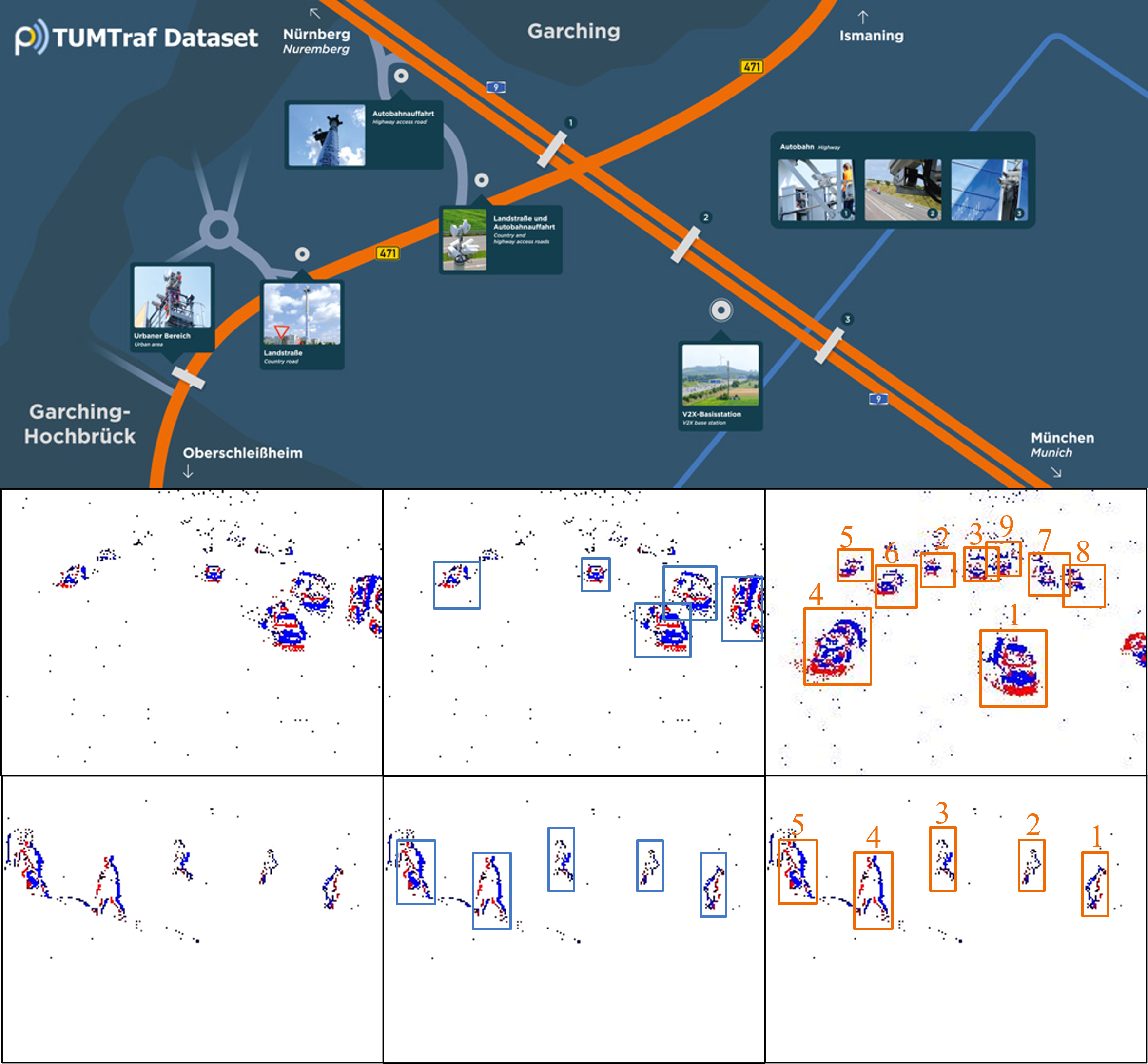}
    \caption{This figure shows the project of TUMTraf, all the data are collected on the roadside of A9 in Garching by Munich. 
    For more details on the dataset, visit \href{https://innovation-mobility.com/en/project-providentia/a9-dataset/}{Project Providentia++ A9 Dataset}. 
    The second and third rows of our dataset display a sequence from left to right: the original image, a visualization of the detection and the tracking.}
    \label{fig:TUMTraf}
\end{figure}

Although event-camera research has made remarkable progress in recent years, roadside ITS still lacks specific, well-labeled MOT data and a commonly accepted evaluation protocol. This is due to two factors: the shortage of annotated event-based datasets and the limited development of event-based tracking methods~\cite{Lagorce2017HOTSAH}. Annotating event streams is costly, as events arrive asynchronously at microsecond resolution and lack texture, making bounding-box and identity labeling labor-intensive. Most studies on event cameras for ITS focus on object detection, often assuming that better detection will naturally translate into better tracking. Yet, tracking itself has received comparatively limited attention.

To address this gap, we release a dataset focused on tracking pedestrians and vehicles, targeting applications in autonomous driving and road surveillance. The sequences span daytime and evening traffic, varying densities, with time-synchronized labels for vehicles and pedestrians. Recent efforts have explored event-based object detection using neural networks~\cite{He2021FASTDynamicVisionDA} and clustering algorithms~\cite{Chen2018NeuromorphicVB}, but few studies target traffic object tracking. 
\cite{Chen2018NeuromorphicVB} introduced detection and tracking by clustering with probabilistic filters. Its follow-up, \cite{Chen2019MultiCueEI}, incorporated CNN-based detection to enhance accuracy, but also revealed that sudden changes in pedestrian motion and high object density substantially degrade online tracking-by-clustering. Nevertheless, a benchmark that jointly evaluates event-based detection and tracking under standardized protocols and challenging real-world scenarios is still lacking. Thus, we establish a tracking-by-detection benchmark for consistent evaluation in ITS.

The TUMTraf real-world dataset series (see Fig.~\ref{fig:TUMTraf}) integrates RGB cameras, Doppler radars, LiDARs, and event-based cameras. It encompasses diverse traffic and weather-related scenarios. One dataset highlights critical traffic and crash situations on the A9, using event-based imaging under daytime and evening conditions to advance cooperative 3D object detection and tracking\cite{Cre2024TUMTrafEC}. As another branch of the TUMTraf series, TUMTraf EMOT aims to provide event-based data with tracking labels, enabling integrated detection-and-tracking research.

Our main contributions can be summarized as follows:

\begin{itemize}
\item We introduce the TUMTraf EMOT dataset, specifically designed for Intelligent Transportation Systems (ITS), which includes unique video sequences for the separate tracking of vehicles and pedestrians.
\item We present a standardized evaluation protocol and strong tracking-by-detection baselines for event data, along with pre-trained models, enabling a practical trade-off between efficiency and accuracy.
\item We evaluate our model that shows outstanding results on the TUMTraf EMOT dataset, establishing a robust benchmark for subsequent studies in this field.
\end{itemize}

\section{Related Work}

\subsection{Event-based Dataset}

\begin{table}[!t]
\centering
\caption{The summary for various event-based datasets.}
\resizebox{\linewidth}{!}{ 
\begin{tabular}{c|ccccc}
\toprule[1.5 pt]
Dataset & Class  & Boxes & Label & Detection & Tracking \\ 
\midrule
Gen1~\cite{Tournemire2020ALS} & 2  & 255K & Pseudo  & \checkmark & $\times$ \\ 
1MegaPixel~\cite{Perot2020LearningTD} & 3  & 25M & Pseudo & \checkmark & $\times$ \\ 
TUMTraf Event~\cite{Cre2024TUMTrafEC} & 7  & 4k & Pseudo  & \checkmark & $\times$ \\ 
PKU-DAVIS-SOD~\cite{Li2023SODFormerSO} & 3  & 1.1M & Manual& \checkmark & $\times$ \\ 
\midrule
Ours & 2 & 54M & Manual  & \checkmark & \checkmark \\ 
\bottomrule[1.5pt]
\end{tabular}
}
\label{table:dataset_det}
\end{table}

\begin{figure}[!t]
    \centering
    \includegraphics[width=0.9\linewidth]{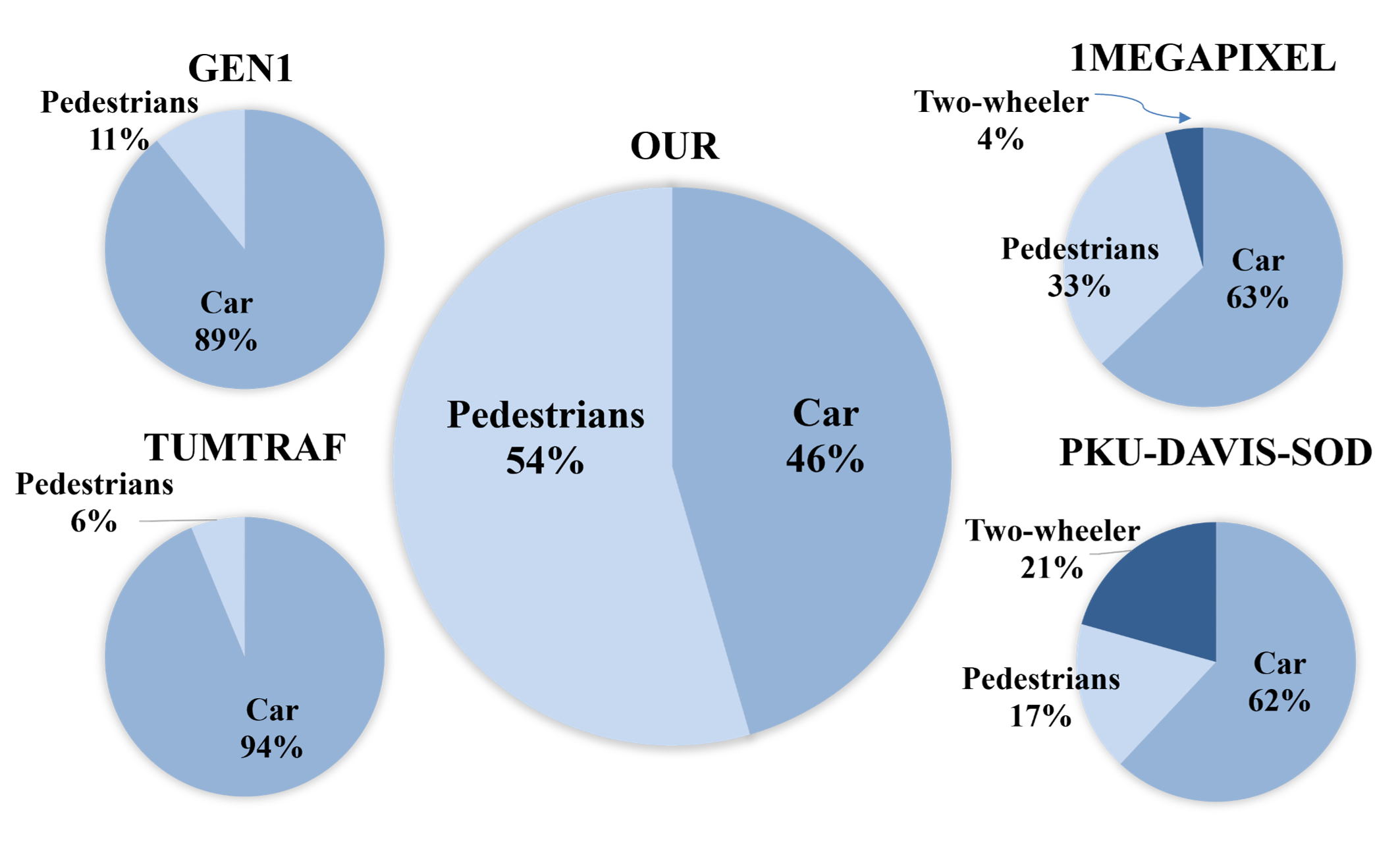}
    \caption{This figure shows the percentage of each class in their label numbers among several event-based datasets.}
    \label{fig:dataset_all}
\end{figure}
We review public event-based datasets for detection and tracking for ITS. Table~\ref{table:dataset_det} highlights that current datasets primarily provide detection annotations without unified ground truth for tracking-by-detection. 

Among the pioneering datasets in this domain, the DDD17~\cite{Binas2017DDD17ED} dataset offers 12 hours of annotated recordings from a DAVIS sensor, capturing diverse driving scenarios and crucial vehicle parameters. However, it was limited by its scope regarding road types, weather, and lighting conditions. To address this, the DDD20~\cite{Hu2020DDD20EE} dataset extended DDD17 with an additional 39 hours of data, including glare environments, urban driving, and mountain highway driving, and increases the data volume to 1.3 TB. 

Moreover, the Gen1 Detection dataset~\cite{Tournemire2020ALS} comprises over 39 hours of automotive recordings from a PROPHESEE GEN1 sensor. This dataset spans a wide array of driving scenarios, from urban and highway scenes to suburban and countryside settings, under diverse weather and lighting conditions. It includes manual bounding box annotations for cars and pedestrians provided at frequencies between 1 and 4 Hz, resulting in over 255,000 labels. In~\cite{Perot2020LearningTD}, the 1Mpx automotive detection dataset recorded with a high-resolution (1280$\times$720) event camera, and proposed a fully automated labeling protocol. The dataset provides over 25 million bounding boxes for cars, pedestrians, and two-wheelers at 60 Hz.

Additionally, TUMTraf Event~\cite{Cre2024TUMTrafEC} also recorded data at different locations along the same road segment with us, comprising 93,000 images. This dataset includes annotations for various types of vehicles, such as trucks and trailers. Smaller-scale datasets like PKU-DAVIS-SOD~\cite{Li2023SODFormerSO} offer 220 roadside sequences and a total of 1,080,100 bounding boxes. The dataset separate different categories like: cars, pedestrians, and two-wheelers, with proportions in each subset being roughly 3.5:1:1.2, 6:1:1, and 3.5:1.5:1, respectively.

\subsection{Tracking-by-detection methods for event camera}

Tracking-by-detection has been widely adopted in MOT for camera imagery, but its use with event-based data remains limited. Early works primarily relied on unsupervised clustering methods. For example, GSCEventMOD~\cite{Mondal2021MovingOD} applies graph spectral clustering, while classical approaches such as MeanShift~\cite{Comaniciu2002MeanSA}, DBSCAN~\cite{Ester1996ADA}, and WaveCluster~\cite{Sheikholeslami1998WaveClusterAM} detect cluster centers or dense regions in the event stream. These methods treat cluster centers as detection outputs, but they struggle under high object density and complex motion. On the supervised learning aspect, \cite{Perot2020LearningTD} proposed dual regression heads with temporal consistency loss to enhance detection accuracy. In parallel, general detectors such as YOLO~\cite{Redmon2015YouOL} and its successor YOLOX~\cite{Ge2021YOLOXEY} can be adapted to event camera data, which are already demonstrated fast and accurate detection in frame-based image. More recently,  event–frame fusion techniques with attention modules~\cite{9546775,cao2024embracing} have been explored, achieving further performance gains by leveraging complementary information from both modalities.

In tracking-by-detection, early research largely relied on probabilistic models. Bayesian filters, particularly the GM-PHD filter~\cite{Vo2006TheGM}, estimated both object states and cardinality by modeling birth, survival, and detection probabilities. Extensions combined probabilistic data association with deep learning detectors and appearance cues~\cite{BarShalom2009ThePD,Musicki2002JointIP}, but these approaches were still limited by single-point hypothesis assumptions. Then the emphasis has shifted toward lightweight, real-time trackers. SORT~\cite{Bewley2016SimpleOA} demonstrated the efficiency of combining Kalman filtering with Hungarian-based association, though it was prone to ID switches and trajectory fragmentation. DeepSORT~\cite{Wojke2017SimpleOA} addressed these shortcomings with re-identification features, while ByteTrack~\cite{Zhang2021ByteTrackMT} improved association by leveraging low-confidence detections. Building on these advances, newer variants such as StrongSORT~\cite{Du2022StrongSORTMD}, BoT-SORT~\cite{Aharon2022BoTSORTRA}, and Hybrid-SORT~\cite{Yang2023HybridSORTWC} have focused on robust occlusion handling and identity preservation. These real-time methods not only validate the effectiveness of online tracking, but also provide transferable frameworks for event-based data, as their data association relies on generic detection results rather than modality-specific features.

For event data in ITS, the authors of~\cite{Chen2018NeuromorphicVB} built the first benchmark and they also explored different trackers, such as GM-PHD~\cite{Vo2006TheGM}, Probabilistic Data Association (PDA)\cite{BarShalom2009ThePD}, even through SORT\cite{Bewley2016SimpleOA} method. These methods have been applied to vehicle and pedestrian tracking. Although effective, they still struggle to distinguish between different objects in environments with rapidly changing object densities or high levels of clutter.

\section{TUMTraf EMOT: Event-based Multi-Object Tracking Dataset}

\begin{table}[t!]
\centering
\caption{Dataset description for vehicle tracking. It includes three event sequences.}
\begin{tabular}{c|ccc}
\toprule[1.5pt]
\textbf{Events} & \textbf{Duration (s)} & \textbf{Events (M)} & \textbf{Average (Keps)} \\ 
\midrule
EventSeq-Vehicle1 & 45.4 & 110.7 & 2438 \\ 
EventSeq-Vehicle2 & 32.4 & 79.4 & 2450 \\ 
EventSeq-Vehicle3 & 21.8 & 53.4 & 2450 \\ 
\bottomrule[1.5pt]
\end{tabular}
\label{table:vehicle_set}
\end{table}

\begin{table}[t!]
\centering
\caption{Dataset description for pedestrian tracking. It includes three event sequences.}
\begin{tabular}{c|ccc}
\toprule[1.5pt]
\textbf{Events} & \textbf{Duration (s)} & \textbf{Events (M)} & \textbf{Average (Keps)} \\
\midrule
EventSeq-Pedestrian1 & 14.4 & 34.3 & 2382 \\
EventSeq-Pedestrian2 & 38.2 & 89.6 & 2345 \\
EventSeq-Pedestrian3 & 33.2 & 88.9 & 2677 \\ 
\bottomrule[1.5pt]
\end{tabular}
\label{table:pedestrian_set}
\end{table}

In this section, we introduce the TUMTraf EMOT dataset, an event-based dataset designed for roadside surveillance. As shown in Fig.\ref{fig:dataset_all}, existing datasets are largely vehicle-focused, with relatively few pedestrian samples, resulting in a clear class imbalance. In contrast, TUMTraf EMOT addresses these limitations by offering a balanced dataset with integrated detection and tracking labels.

\subsection{Event camera}
\begin{figure*}[t!]
    \centering
    \includegraphics[width=0.8\linewidth]{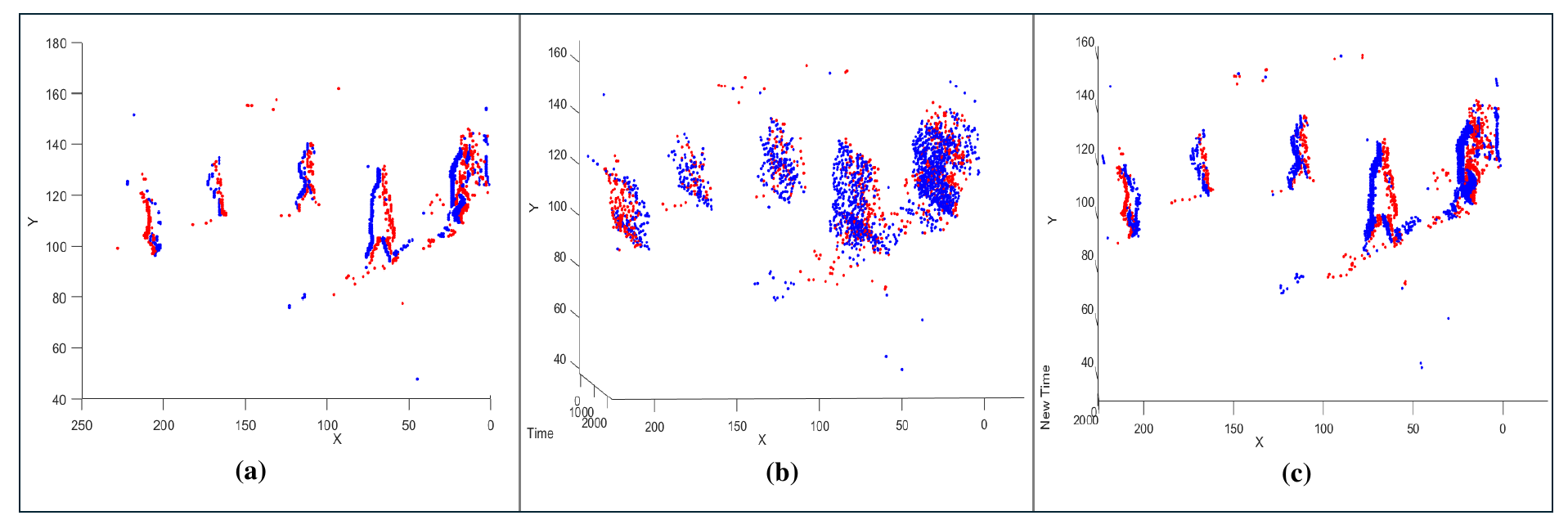}
    \caption{The illustration of how events are gathered into a framed image. Figure (a) shows the event points at an earlier instant, Figure (c) at a later instant, and Figure (b) illustrates the accumulation of events between these two instants. Points are colored by event polarity to indicate motion: red = brightness increase (positive), blue = brightness decrease (negative).}
    \label{fig:events_camera}
\end{figure*}

Event cameras represent a departure from traditional frame-based cameras by asynchronously detecting pixel-level intensity changes~\cite{Chen2020EventBasedNV}, resulting in a continuous stream of events rather than fixed frames. The formula describing the operation of an event camera can be expressed as:

\begin{equation}
\Delta L = L(x, y, t) - L(x, y, t - \Delta t) \geq \theta 
\end{equation}
where \(L(x, y, t)\) represents the log intensity at pixel \((x, y)\) and time \(t\), while \(\Delta L\) denotes the change in log intensity over time $\Delta t$. Events are triggered when this change $\Delta L$ exceeds a predefined contrast threshold $\theta$, facilitating the capture of fine temporal details (see Fig. \ref{fig:events_camera}).

For convolutional processing, events are usually accumulated over a short time window~$\Delta t$ and projected onto a fixed 2D grid as polarity count images. This conversion introduces two main effects: (i)\emph{Reduced effective spatial resolution}: only motion edges are recorded, leaving object interiors largely empty, and far-field objects produce very few active pixels. Temporal accumulation further discretizes these sparse edges into grid cells, lowering the signal-to-noise ratio and making fine details occupy only a handful of pixels. (ii)\emph{Crowding of small objects}: within the same $\Delta t$, events from nearby objects (e.g., vehicles in traffic) often overlap or cluster in adjacent cells. Occlusions and parallax can merge traces, creating dense, ambiguous patterns that local receptive fields struggle to separate.

\subsection{Data collection}

\begin{figure}[!t]
    \centering
    \includegraphics[width=0.9\linewidth]{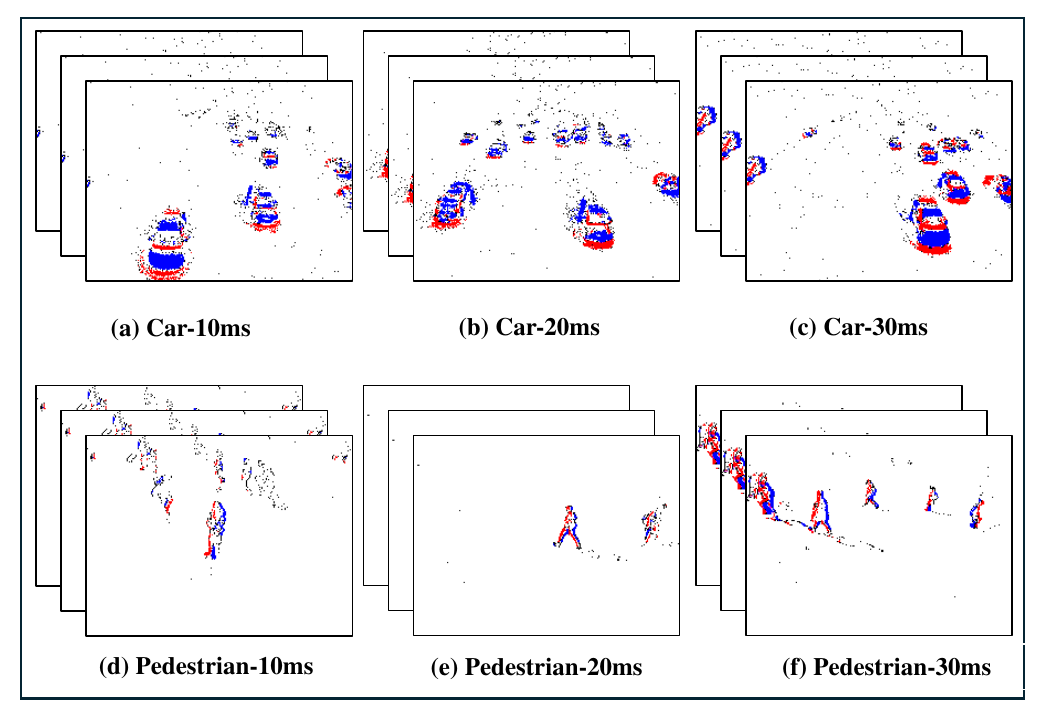}
    \caption{The visualization of event sequences at different time intervals. The dataset consists of six distinct event sequences, each characterized by intervals of 10 ms, 20 ms, and 30 ms.}
    \label{fig:datasets}
\end{figure}

The dataset, intended for object tracking, is primarily oriented towards vehicle and pedestrian surveillance scenarios. Data collection utilized the DAVIS240 (Dynamic and Active Pixel Sensor-DAVIS), which offers a resolution of 240$\times$180 pixels. Our research emphasizes event-based processing for object detection and tracking, facilitating benchmark comparisons across different methodologies using event data~\cite{Chen2018NeuromorphicVB}. Our dataset is divided into six distinct event sequences (see Fig.~\ref{fig:datasets}), categorized based on varying time intervals and specific object attributes, namely cars and pedestrians. For each object group, event sequences were constructed with time intervals of 10 ms, 20 ms, and 30 ms, facilitating analysis of vehicle and pedestrian movements within different temporal scales.

We divide the dataset into sequences with three time intervals(10 ms, 20 ms, 30 ms) because temporal aggregation strongly affects event frame characteristics. Short intervals (e.g., 10 ms) yield higher noise and incomplete object shapes, while long intervals (e.g., 30 ms) cause motion streaks\cite{Gallego2018AUC} and ID Switching. In traffic scenes, dense flows require shorter intervals to separate overlapping objects, whereas sparse scenes benefit from longer intervals to reduce noise and computation. Moreover, the chosen values cover typical DAVIS240 event densities, span high- to low-speed motions, and make results comparable to 30 FPS frame-based baselines.

\fakeparagraph{Dataset for vehicle tracking.}
Vehicle tracking data was acquired using the DAVIS240 positioned on a bridge. The dataset comprises three distinct event sequences detailed in Table~\ref{table:vehicle_set}. Vehicles in these sequences are observed traveling in both directions, approaching and moving away from the camera across multiple lanes. This dataset encompasses a broad spectrum of vehicle types, ranging from small cars to large trucks, which significantly enriches its diversity and complexity.

\fakeparagraph{Dataset for pedestrian tracking.}
For pedestrian tracking, data was collected from three distinct scenarios, as detailed in Table~\ref{table:pedestrian_set}. The first event sequence involved surveillance near a traffic signal, resulting in a 14.4-second sequence containing 34.3 million events, which highlighted challenges in dense urban environments. The second event sequence focused on outdoor pedestrian tracking near a subway station, yielding a 38.2 second sequence with 89.6 million events, showcasing varied pedestrian densities and movement patterns. The third event sequence captured an indoor scenario with pedestrians moving in different directions, producing a 33.2 second sequence with 88.9 million events. 

\subsection{Dataset annotation}

The dataset was manually annotated using the ViTBAT tool~\cite{Biresaw2016ViTBATVT}. The annotated data is provided in .mat and .txt formats, prepared specifically for training and testing. To align with the standards set by the MOT Challenge~\cite{Milan2016MOT16AB}, we formatted ViTBAT's output into a .json structure identical to the ground truth file of the MOT16 Challenge. This conversion ensures compatibility with widely used multi-object tracking benchmarks, thus improving the usability of the dataset for researchers and developers.

\subsection{Dataset release}
We release the raw event streams in the original sensor format, together with the event-image sequences used in our experiments. Specifically, we provide event images generated using fixed time intervals of 10 ms, 20 ms, and 30 ms. In addition, we also include the code for generating the event-image sequences from the raw event data, along with the official train/test splits.

\section{Baseline Method}

\begin{figure*}[!t]
    \centering
    \includegraphics[width=1\linewidth]{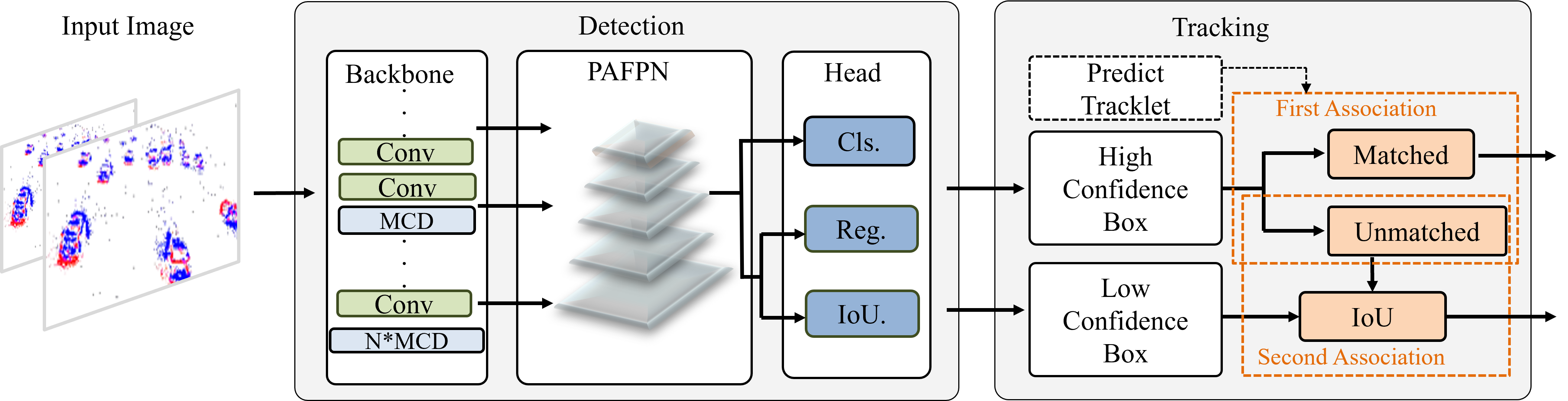}
    \caption{The architecture of our tracking-by-detection framework.}
    \label{fig:our_pipeline}
\end{figure*}

This section introduces an efficient event-based tracking-by-detection baseline. The entire pipeline is illustrated in Fig. ~\ref{fig:our_pipeline}, which comprises event-based object detection and multi-object tracking parts.

\subsection{Event-based object detection}
\fakeparagraph{Overview.}
As shown in Fig.~\ref{fig:our_pipeline}, the architecture of the event-based object detection network consists of a backbone, a Path Aggregation Feature Pyramid Network (PAFPN)\cite{liu2018path}, and decoupled heads. Specifically, we incorporated MCD resblocks into the backbone network to extract more robust feature representations. Similar to \cite{Ge2021YOLOXEY}, we use PAFPN for multi-level feature fusion and decoupled heads for detection predictions. The computation process can be represented as follows:

\begin{equation}
\begin{aligned}
C_{k} &=F_{\text {backbone }}(I) \\
P_{k} &=F_{\text {pafpn }}\left(C_{k}\right) \\
Y_{k} &=F_{\text {head }}\left(P_{k}\right), k=3,4,5
\end{aligned}
\end{equation}
where $I$ is the input event image and $\left\{C_{k}\right\}_{k=3}^{5}$ are the extracted multi-level features. $\left\{P_{k}\right\}_{k=3}^{5}$ and $\left\{Y_{k}\right\}_{k=3}^{5}$ represent the fused multi-scale features and the corresponding prediction outputs, respectively. $F_{\text {backbone}}$, $F_{\text {pafpn}}$, and $F_{\text {head}}$ denote the backbone function, the PAFPN function, and the decoupled heads function, respectively. We elaborate on each of these components below.

\begin{figure}[!t]   
\centering
    \includegraphics[width=1\linewidth]{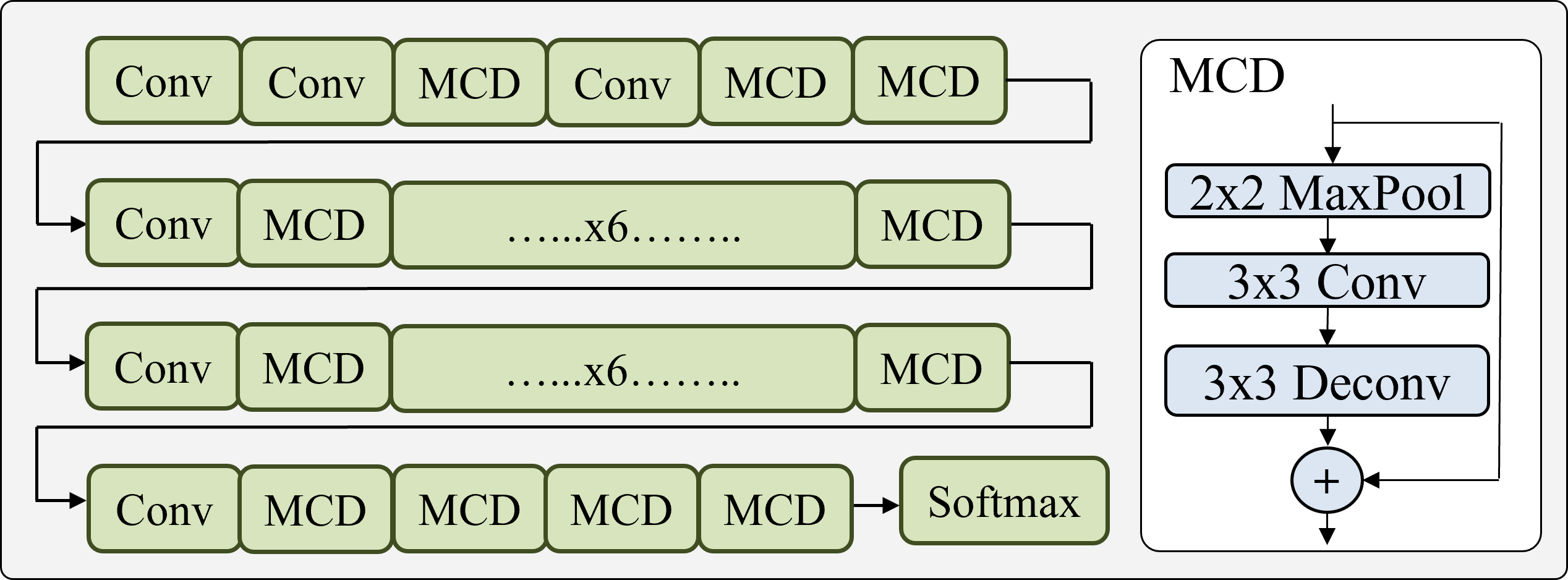}
    \caption{The structure of the backbone network, \texttt{MCD} denotes our new feature extractor, which replaces residual blocks with coupled convolution–deconvolution layers to improve feature extraction from low-resolution event data.}
    \label{fig:extractor}
\end{figure}

\fakeparagraph{Backbone.}
Darknet-53\cite{Redmon2018YOLOv3AI} is used as the backbone network to extract feature representations. It is distinguished by its 53 convolutional neural network layers architecture, each followed by batch normalization and the Leaky ReLU activation function. Notably, the backbone incorporates residual connections to enable the training of deeper networks without performance degradation, supporting the extraction of higher-level semantic features. 
Our work introduces a modified bottleneck structure, termed MCD resblocks to address both mentioned issues from event-based images: (i)~\emph{Low effective spatial resolution} and (ii)~\emph{Crowding of small objects}. As illustrated in Fig.~\ref{fig:extractor}, MCD resblocks are designed to extract features by briefly moving computation to a coarser scale to gather context (mitigating sparsity and ambiguity), and then restoring the original resolution while preserving high-frequency detail through a residual path. The detailed process is as follows:
\begin{equation}
\begin{aligned}
 M(x) &= \text{MaxPool2d}(x)  \\
 C(x) &= \text{LeReLU}(\text{Conv}(M(x)))  \\
 T(x) &= \text{LeReLU}(\text{BN}(\text{ConvTranspose2d}(C(x)))) \\
 Y &= x + T(x) 
\label{eq:output}
\end{aligned}
\end{equation}
Where $x$ denotes the input.

We first apply \texttt{MaxPool2d} to aggregate motion responses over local neighborhoods, retaining the strongest evidence and enlarging the receptive field at low cost. It then passes to a \(3\times3\) convolution with LeakyReLU for learnable, context-aware filtering at a coarser scale. When several small objects are only a few pixels apart (sometimes within the same grid cell at 1/8 or 1/16 resolution), a standard \(3\times3\) field may confuse one large object with multiple instances. We therefore add a branch with a larger effective receptive field, providing group-level context so the head can decide instance membership and whether it is a target. Next, \texttt{ConvTranspose2d}+BatchNorm+LeakyReLU performs learnable upsampling, projecting coarse-scale context back to the native grid to reconstruct context-consistent, sharpened activations rather than naive interpolation. Finally, a skip connection forms \(Y = x + T(x)\), preserving high-frequency edges from \(x\) while adding disambiguating context \(T(x)\). In combination, these steps help extract information from sparse, low-resolution event frames and mitigate small-object crowding in dense highway scenes.

\fakeparagraph{PAFPN.}
The PAFPN network further integrates the extracted features $\left\{P_{k}\right\}_{k=3}^{5}$ from different layers. Building on the Feature Pyramid Network (FPN)~\cite{Lin2016FeaturePN}, which uses a top-down approach to create high-level semantic feature maps at various scales, PAFPN utilizes an additional bottom-up path from PANet~\cite{liu2018path} to propagate strong low-level features upwards. This dual-pathway aggregation ensures a rich and balanced flow of information, improving detection accuracy, especially for small and medium-sized objects. 

\fakeparagraph{Decoupled heads.}
All features from PAFPN are sent to decoupled heads for the final task: classification, regression, and IoU. In a decoupled head, parameters are not shared between different predicted feature maps. Given feature maps from the PAFPN at different levels, denoted by P3, P4, and P5 with dimensions \(H \times W \times C \), the decoupled head operations can be described as follows:
\begin{equation}
\begin{aligned}
H_{\text{cls}}(P_k) &= \text{Softmax}( \text{Conv}_{1 \times 1}(\text{Conv}_{3\times3}(\text{Conv}_{1\times1}(P_k))))  \\
H_{\text{reg}}(P_k) &= \text{Conv}_{1 \times 1}( \text{Conv}_{3\times3}(\text{Conv}_{1\times1}(P_k))) \\
H_{\text{iou}}(P_k) &= \sigma( \text{Conv}_{1 \times 1}(\text{Conv}_{3\times3}(\text{Conv}_{1\times1}(P_k)))) 
\label{eq:heads}
\end{aligned}
\end{equation}
where \( P \) represents the feature maps from the PAFPN, \( Conv_{1\times1} \) and \( Conv_{3\times3} \) denote $1\times 1$ and $3\times 3$ convolutions, respectively, \( H_{\text{cls}} \), \( H_{\text{reg}} \), and \( H_{\text{iou}} \) represent decoupled heads for classification, regression, and objectness, respectively. Finally, \( \sigma \) represents the sigmoid function used for objectness prediction.

\subsection{Multi-object tracking}
\begin{figure}[!t]     
\centering
    \includegraphics[width=\linewidth]{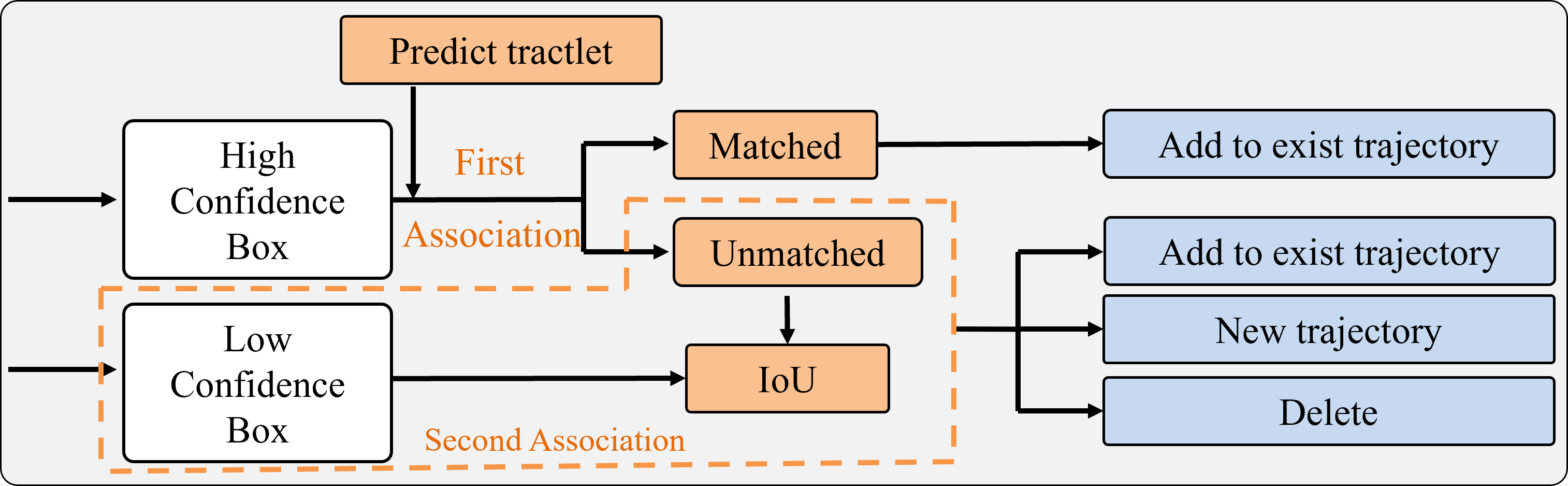}
    \caption{The multi-object tracking pipeline operates as follows. Predicted tracklets are categorized into high- and low-confidence boxes. First, high-confidence boxes are matched to existing trajectories. Second, IoU-based association links the remaining unmatched trajectories with low-confidence boxes. Matched boxes update existing trajectories or initialize new ones; unmatched boxes are discarded.}
    \label{fig:tracker}
\end{figure}

For the tracking part, we employ a single hypothesis tracking strategy (see Fig.~\ref{fig:tracker}), incorporating a standard Kalman filter for state estimation and utilizing the Hungarian method for data association\cite{Bewley2016SimpleOA}. The process is as follows: Detected boxes are divided into two sets—higher-score and lower-score boxes. Trajectories are also categorized as active and inactive. An active trajectory is one that has tracked the target for more than two frames, including any new trajectories that start in the initial frame.

\begin{algorithm}[!t]
\caption{Pseudo-code of our tracker using mathematical notation}
\label{alg:bytetrack}
\begin{algorithmic}[1]
\State \textbf{Input:} A video sequence $V$, object detector Det, detection score threshold $\tau$
\State \textbf{Output:} Tracks $T$ of the video
\State Initialization: $T \leftarrow \emptyset$
\For{$f$ in $V$} \tikzmark{start}
    \State $D_f \leftarrow \text{Det}(f)$
    \Statex \hfill \Comment{Predict detection boxes \& scores}
    \State $D_{\text{high}} \leftarrow \{d \in D_f : \text{score}(d) > \tau \}$
    \State $D_{\text{low}} \leftarrow \{d \in D_f : \text{score}(d) \leq \tau \}$
    \State $T' \leftarrow \{\text{KalmanFilter}(t) : t \in T \}$
    \Statex \hfill \Comment{Estimate the new state}
    \State $(T_{\text{remain}}, D_{\text{high\_remain}}) \leftarrow \text{Associate}(T', D_{\text{high}}, C_1)$
    \Statex \hfill \Comment{First association}
    \State $(T_{\text{remain\_final}}, D_{\text{low\_remain}}) \mspace{-6mu} \leftarrow \mspace{-6mu} \text{Associate}(T_{\text{remain}}, D_{\text{low}}, C_2)$
    \Statex \hfill \Comment{Second association}
    \State $T \leftarrow T \setminus T_{\text{remain\_final}}$
    \State $T \leftarrow T \cup \{\text{InitializeTrack}(d) : d \in D_{\text{high\_remain}} \cup D_{\text{low\_remain}} \}$
    \Statex \hfill \Comment{Initialize new tracks}
\EndFor \tikzmark{end}
\State \textbf{return} $T$
\end{algorithmic}
\end{algorithm}

\fakeparagraph{First association.}
For high-score bounding boxes, correlating them with previous trajectories is straightforward. Similarity is computed using Intersection over Union (IoU), and the prediction result from the Kalman filter.

Based on the similarity, the initial tracking trajectory and the current frame's high-score bounding box are matched using Eq.~\ref{eq:hungarian} (the Hungarian algorithm). This results in three outcomes: matched trajectories and bounding boxes, trajectories that were not successfully matched, and bounding boxes from the current frame that were not successfully matched. The cost of assigning detection \( i \) to track \( j \) at time \( t \) is represented by the cost matrix \( C \) where each element \( C_{ij} \) is calculated as follows:

\begin{equation}
    C_{ij} = \text{cost}(D_{t_i}, T_{t_j}) \label{eq:hungarian}
\end{equation}
where \( C_{ij} \) is the cost of assigning detection \( i \) to track \( j \), \( D_{t_i} \) is the \( i \)-th detection at time \( t \), and \( T_{t_j} \) is the \( j \)-th track at time \( t \). The Hungarian algorithm is applied to the cost matrix \(C \) to find the optimal assignment of detections to tracks, minimizing the total assignment cost. After association, the preliminary tracking trajectory is updated using the current frame's successfully matched bounding box, maintaining the original ID, as shown in Algorithm~\ref{alg:bytetrack}.

\fakeparagraph{Second association.}
For low-confidence bounding boxes, IoU is computed with trajectories unmatched after the initial association and use the Hungarian algorithm to pair them in the current frame. Update the matched trajectories with the current-frame boxes. Exclude trajectories already classified as lost. 
Identify unmatched high-confidence bounding boxes in the current frame and any inactive trajectories. Compute IoU between these trajectories and the unassociated boxes, and use the Hungarian algorithm for matching. If matched, update the corresponding trajectory. Remove any inactive trajectories that remain unmatched.

\fakeparagraph{Object movement.}
Our methodology begins by applying the ByteTrack algorithm~\cite{Zhang2021ByteTrackMT}. Tracking uses two key variables: the object's velocity and direction. Specifically, we add the Hybrid-SORT strategy~\cite{Yang2023HybridSORTWC}. Hybrid-SORT fits event-camera frames better by elevating motion to a primary cue rather than overlap alone. Neuromorphic sensors yield edge-driven, “thin” accumulations; their boxes lower cross-frame IoU, destabilizing matching. Hybrid-SORT injects explicit motion cues—four-corner velocity consistency, a K-frame history, and an inertia prior—into the Hungarian cost, yielding anisotropic, motion-aligned gating that pulls on-trajectory detections closer and pushes clutter away. For fast, non-linear motion typical of event streams, the previous-observation rematch often outperforms pure Kalman extrapolation.
\section{Experimental Settings}
\subsection{Implementation details}
We divided the dataset into six subsets based on three distinct temporal intervals (10 ms, 20 ms, and 30 ms) and two object classes (vehicles and pedestrians). Each subset comprised three event sequences. For clear comparison purposes, we split the entire dataset into equal parts for training and testing. The division was made based on the original sequence of frames, using the first half (from frame 0 to the midpoint) as the training set and assigning the remaining frames of each video sequence to the test set to ensure that all test data remained unseen before inference. To tailor our approach, we processed the dataset at separate time intervals resulting in three distinct sets of pre-trained weights for subsequent inference.

\subsection{Evaluation metrics}

\fakeparagraph{Object detection.}
Precision is the percentage of predictions that are correct, indicating how reliable a model's positive predictions are. Recall measures how well the model can find all the positives, reflecting its ability to identify all relevant instances. These two metrics are typical for understanding the effectiveness of a detection model in various conditions. Their mathematical definitions of \emph{Precision and Recall} are written as:

\begin{equation}
	\label{metrics}
	\begin{aligned}
        \mathrm{Precision}  &= \frac{\mathrm{TP}}{\mathrm{TP} + \mathrm{FP}},\\
		\mathrm{Recall} &= \frac{\mathrm{TP}}{\mathrm{TP} + \mathrm{FN}}.\\
	\end{aligned}
\end{equation}
where TP, FP, and FN denote true positive, false positive, and false negative, respectively. TP + FN equals the number of ground truths.  

\emph{Intersection over Union (IoU).}
The IoU metric (Eq.~\ref{iou}) is often used as a threshold to determine whether a predicted detection is labeled as a true positive or a false positive. Different IoU thresholds, such as IoU = 0.5 or IoU = 0.75, may be used depending on the specific evaluation criteria. 

\begin{equation}
	\label{iou}
	\begin{aligned}
		\mathrm{IoU} &= \frac{\mathrm{Detections}\cap \mathrm{Groundtruth}}{\mathrm{Detections}\cup \mathrm{Groundtruth}}.\\
	\end{aligned}
\end{equation}

The precision-recall curve is a graphical representation of the trade-off between precision and recall at different classification thresholds. The area under this curve represents the average precision (AP) for a particular class. Taking the mean of all APs across all classes yields the mean average precision (mAP), a common metric for evaluating object detection models.
In addition to mAP, other metrics based on different IoU thresholds and object sizes are often used to provide a more detailed evaluation:

\begin{itemize}
    \item $\mathrm{AP_{50}}$ (AP at IoU = 0.5): This metric measures precision when there is at least a $50\%$ overlap between predicted and ground truth bounding boxes.
    \item $\mathrm{AP_{75}}$ (AP at IoU = 0.75): This metric measures precision when there is at least a $75\%$ overlap between predicted and ground truth bounding boxes.
    \item $\mathrm{AP_S}$ (AP for Small Objects): This metric focuses on small-sized objects (e.g., area less than $32^2$ pixels).
    \item $\mathrm{AP_M}$  (AP for Medium Objects): This metric focuses on medium-sized objects (e.g., area between $32^2$ pixels and $96^2$ pixels).
    \item $\mathrm{AP_L}$ (AP for Large Objects): This metric focuses on large-sized objects (e.g., areas greater than $96^2$ pixels).
\end{itemize}

\fakeparagraph{Multi-object tracking.}
For multi-object tracking, we employ the metrics established in~\cite{Milan2016MOT16AB}. The detailed metrics are as follows:

\begin{itemize}
\item $\mathrm{MOTA\uparrow}$ : Multiple Object Tracking Accuracy.
\item $\mathrm{MOTP\uparrow}$ : Multiple Object Tracking Precision.
\item $\mathrm{MT\uparrow}$ : Number of Mostly Tracked Trajectories.
\item $\mathrm{PT\uparrow}$ : Partially Tracked Trajectories.
\item $\mathrm{ML\downarrow}$ : Mostly Lost Trajectories.
\item $\mathrm{FP\downarrow}$ : False Positives.
\item $\mathrm{FN\downarrow}$ : False Negatives.
\item $\mathrm{IDs\downarrow}$ : ID Switches.
\item $\mathrm{IDF1\uparrow}$ : Identification F-Score.
\item $\mathrm{IDP\uparrow}$ : Identification Precision.
\item $\mathrm{IDR\uparrow}$ : Identification Recall.
\end{itemize}

Higher scores for metrics denoted by ("↑") indicate better performance, while lower scores for metrics denoted by ("↓") also indicate better performance, as shown in Section~\ref{results}. \\

MOTA provides an overall assessment of tracking performance by considering both detection and tracking errors. MOTP calculates the average distance between the predicted and ground truth object positions across all frames, offering a measure of the tracking algorithm's precision. MOTA and MOTP can be computed as follows:

\begin{equation}
\label{MOTA}
\begin{aligned}
\text{MOTA} &= 1 - \frac{\sum_t (\text{FN}_t + \text{FP}_t + \text{IDs}_t)}{\sum_t \text{GT}_t} \\
\text{MOTP} &= \frac{\sum_{k=1}^{N} \sum_{i=1}^{M_k} d_{i,k}}{\sum_{k=1}^{N} M_k} 
\end{aligned}
\end{equation}
where IDs refer to ID switches, GT indicates ground truths, and $t$ denotes the time step. Meanwhile, $N$ is the total number of frames, $M_k$ is the number of true positive detections in frame $k$, and $d_{i,k}$ is the distance between the predicted position and the ground truth position of the $i^{th}$ detection in frame $k$. 

Recognizing the need for comprehensive evaluation, we incorporated additional metrics specific to identity tracking: ID F1 Score (IDF1), ID Recall (IDR), and ID Precision (IDP). IDR evaluates the tracker's ability to consistently recognize identities across frames, while IDP assesses the accuracy of these identity assignments. The IDF1 Score focuses on assessing how well a system maintains consistent identities across frames~\cite{Luiten2020HOTAAH} and emphasizes the importance of accurately tracking the continuity of each individual object throughout a sequence. IDF1 provides a more holistic assessment of tracking performance by explicitly penalizing identity switches and fragmentation, which are often overlooked by traditional metrics.

\begin{equation}
\begin{aligned}
\text{IDF1} &= \frac{2 \times \text{IDTP}}{2 \times \text{IDTP} + \text{IDFP} + \text{IDFN}} \\
    \text{IDP} &= \frac{\text{IDTP}}{\text{IDTP} + \text{IDFP}}  \\
    \text{IDR} &= \frac{\text{IDTP}}{\text{IDTP} + \text{IDFN}} \\
\end{aligned}
\end{equation}
where IDTP is identity true positives, IDFP is identity false positives, and IDFN is identity false negatives.

\section{Results and Analyzing}
\label{results}
\subsection{Experimental results for object detection}
After looking through all the evaluation scores, our model provides a basic comparable score for further research. Due to the detection boxes or areas being smaller than the threshold set for large object evaluation, we removed the results from the $AP_{L}$ and did not include them in Table \ref{tab:vehicle_detection_performance} and Table \ref{tab:person_detection_performance}, as this led to a score of -1.

\fakeparagraph{Vehicle detection.}
Table~\ref{tab:vehicle_detection_performance} provides a comprehensive overview of the performance metrics for vehicle detection. Across time windows, vehicles achieve the best overall precision at 10 ms, while $AP_{75}$ remains comparatively low, reflecting difficulty under stricter IoU; longer windows trade accuracy for temporal smoothing.

\fakeparagraph{Pedestrian detection.}
Table~\ref{tab:person_detection_performance} lists the performance metrics for pedestrian detection. For pedestrians, 20 ms yields the strongest detection (highest $AP_{50}$ and $mAP$), but $AP_{75}$ stays low overall, indicating localization under strict overlap is still challenging. The mAP values fluctuate, starting from 13.5\% at 10ms, increasing to 25.7\% at 20ms, and further to 22.6\% at 30ms, suggesting that increases in processing time do not consistently enhance detection efficacy, underscoring the complexities of pedestrian detection in dynamic settings.

\begin{table}[!t]
\caption{Vehicle Detection Performance Over Time}
\label{tab:vehicle_detection_performance}
\centering
\begin{tabular}{c|c|c|c|c|c|c}
\toprule[1.5pt]
Time & $AP_{50}$ & $AP_{75}$ & $AP_{S}$ & $AP_{M}$ & $mAP$ & $FPS$\\
\midrule
10ms & 83.0\% & 31.9\% & 33.8\% & 64.0\% & 39.2\% & 388\\
20ms & 61.3\% & 18.6\% & 24.2\% & 39.8\% & 26.9\% & 379\\
30ms & 75.1\% & 18.5\% & 26.5\% & 51.5\% & 30.6\% & 347\\
\bottomrule[1.5pt]
\end{tabular}
\end{table}

\begin{table}[!t]
\caption{Pedestrian Detection Performance Over Time}
\label{tab:person_detection_performance}
\centering
\begin{tabular}{c|c|c|c|c|c|c}
\toprule[1.5pt]
Time & $AP_{50}$ & $AP_{75}$ & $AP_{S}$ & $AP_{M}$ & $mAP$ & $FPS$\\
\midrule
10ms & 43.6\% & 3.8\% & 11.9\% & 18.0\% & 13.5\% & 424\\
20ms & 70.0\% & 11.9\% & 21.1\% & 38.8\% & 25.7\% & 455\\
30ms & 66.2\% & 8.0\% & 18.7\% & 33.8\% & 22.6\% & 412\\
\bottomrule[1.5pt]
\end{tabular}
\end{table}

\begin{figure}[!t]
    \centering
    \includegraphics[width=\linewidth]{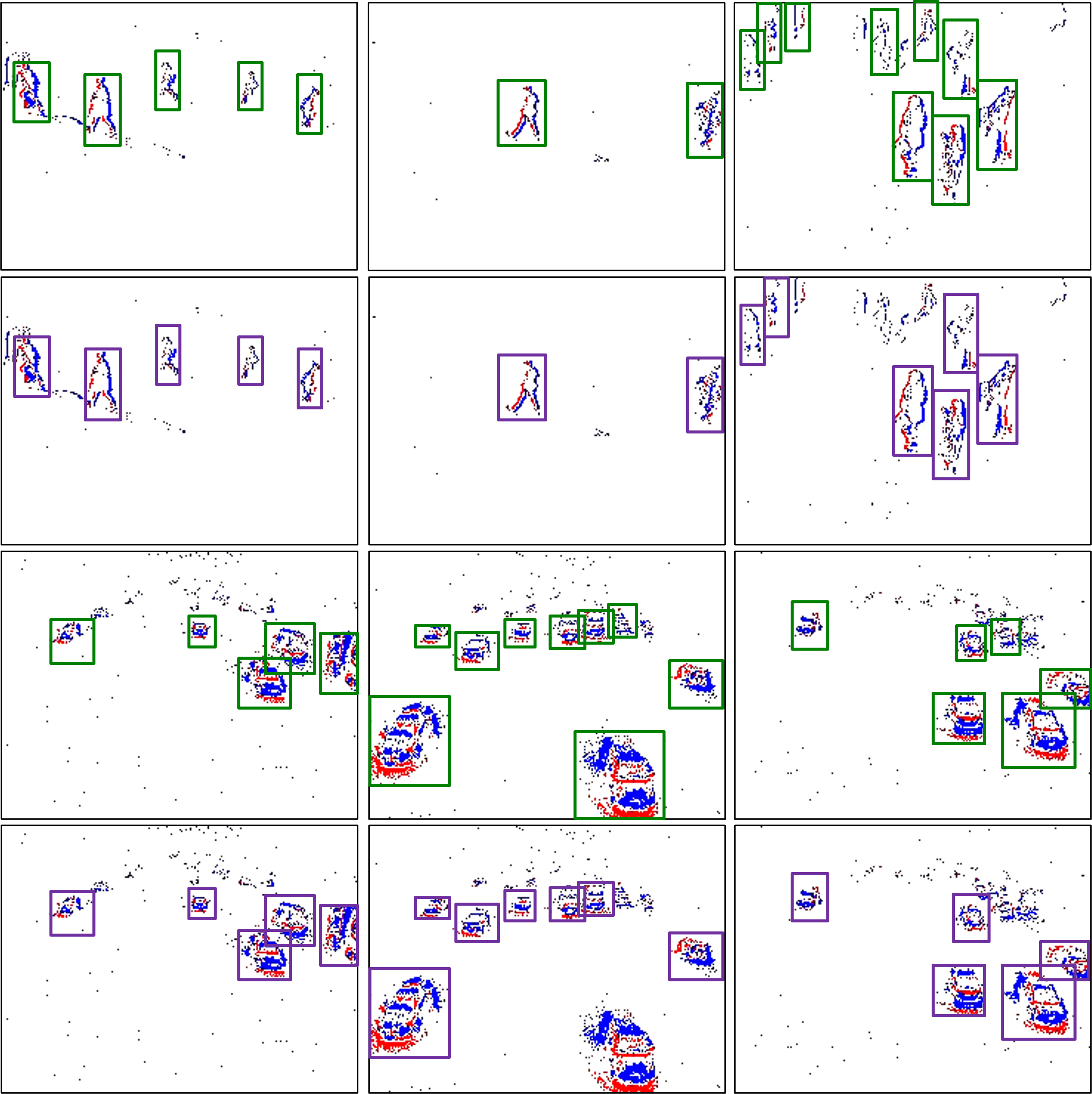}
    \caption{The detection results of vehicles and pedestrians. The first and third rows display the ground truth images, where objects are marked with green boxes. The second and fourth rows show the detection results produced by our method, with objects highlighted in purple boxes.}
    \label{fig:detection visualization}
\end{figure}

\subsection{Experimental results for multi-object tracking}

\fakeparagraph{Vehicle tracking.}
The performance characteristics of the MCD+Hybrid algorithm for vehicle tracking over three different time intervals (10, 20, and 30 ms) are shown in Table~\ref{tab: Vehicle tracking performance}. We report MOTA, MOTP, MT, PT, ML, FP, FN, and IDs as effectiveness indicators. Vehicles track best at 30 ms (highest MOTA/MOTP and fewer IDs), consistent with improved stability from longer temporal aggregation. Compared with 10ms and 20ms, the algorithm reaches its maximum MOTA of 53.5\% at 30ms, indicating higher tracking consistency with fewer identity flips. MOTP improves slightly as the interval increases, reflecting finer localization. Reductions in FP, FN, and IDs indicate better track stability over longer horizons. Identity metrics (IDF1, IDP, IDR) are robust at 10ms and 30ms, reflecting strong alignment between detections and true identities.

\fakeparagraph{Pedestrian tracking.}
The performance of the MCD+Hybrid algorithm in pedestrian tracking shows significant variation across different time intervals in Table \ref{tab: Pedestrian tracking performance}.

\begin{table*}[!t]
\caption{The Performance of Vehicle Tracking at Different Time Intervals}
\label{tab: Vehicle tracking performance}
\centering
\resizebox{\linewidth}{!}{
\begin{tabular*}{\textwidth}{@{\extracolsep{\fill}}c|c|c|c|c|c|c|c|c|c|c|c}
\toprule[1.5pt]
Settings & MOTA↑ & MOTP↑ & MT↑ & PT↑ & ML↓ & FP↓ & FN↓ & IDs↓ & IDF1↑ & IDP↑ & IDR↑\\
\midrule
10ms & 51.1\% &  24.6\% & 26 & 205 & 20 & 2060 & 32384 & 28 & 69.6\% & 93.4\% & 55.5\% \\
\midrule
20ms & 39.3\% & 24.8\% & 29 & 200 & 21 & 4529 & 16851 & 18 & 48.4\% & 57.6\% & 42.6\%\\
\midrule
30ms & 53.5\% & 26.0\% & 29 & 108 & 7 & 829 & 5311 & 10 & 66.5\% & 86.7\% & 53.9\%\\
\bottomrule[1.5pt]
\end{tabular*}
}
\end{table*}

\begin{table*}[!t]
\caption{The Performance of Pedestrian Tracking at Different Time Intervals}
\label{tab: Pedestrian tracking performance}
\centering
\resizebox{\linewidth}{!}{
\begin{tabular*}{\textwidth}{@{\extracolsep{\fill}}c|c|c|c|c|c|c|c|c|c|c|c}
\toprule[1.5pt]
Settings & MOTA↑ & MOTP↑ & MT↑ & PT↑ & ML↓ & FP↓ & FN↓ & IDs↓ & IDF1↑ & IDP↑ & IDR↑\\
\midrule
10ms & 13.0\% & 31.9\% & 3 & 35 & 31 & 1675 & 30026 & 42 & 21.0\% & 34.4\% & 16.0\% \\
\midrule
20ms & 40.0\% & 29.0\% & 17 & 50 & 9 & 4381 & 10407 & 278 & 47.4\% & 58.2\% & 40.1\% \\
\midrule
30ms & 42.1\% & 28.8\% & 13 & 39 & 23 & 1389 & 8247 & 63 & 36.7\% & 44.3\% & 31.7\% \\
\bottomrule[1.5pt]
\end{tabular*}
}
\end{table*}

Despite the relatively low identity matching metrics (IDF1, IDP, IDR) at this interval, which indicate only a fair alignment with actual pedestrian identities, the algorithm manages a lower number of identity switches compared to longer intervals. Pedestrian tracking improves markedly from 10 ms to 20–30 ms; 30 ms attains the highest MOTA, while MOTP and ID metrics suggest a trade-off between localization sharpness and identity consistency.

\begin{figure}[!t]
    \centering
    \includegraphics[width=1\linewidth]{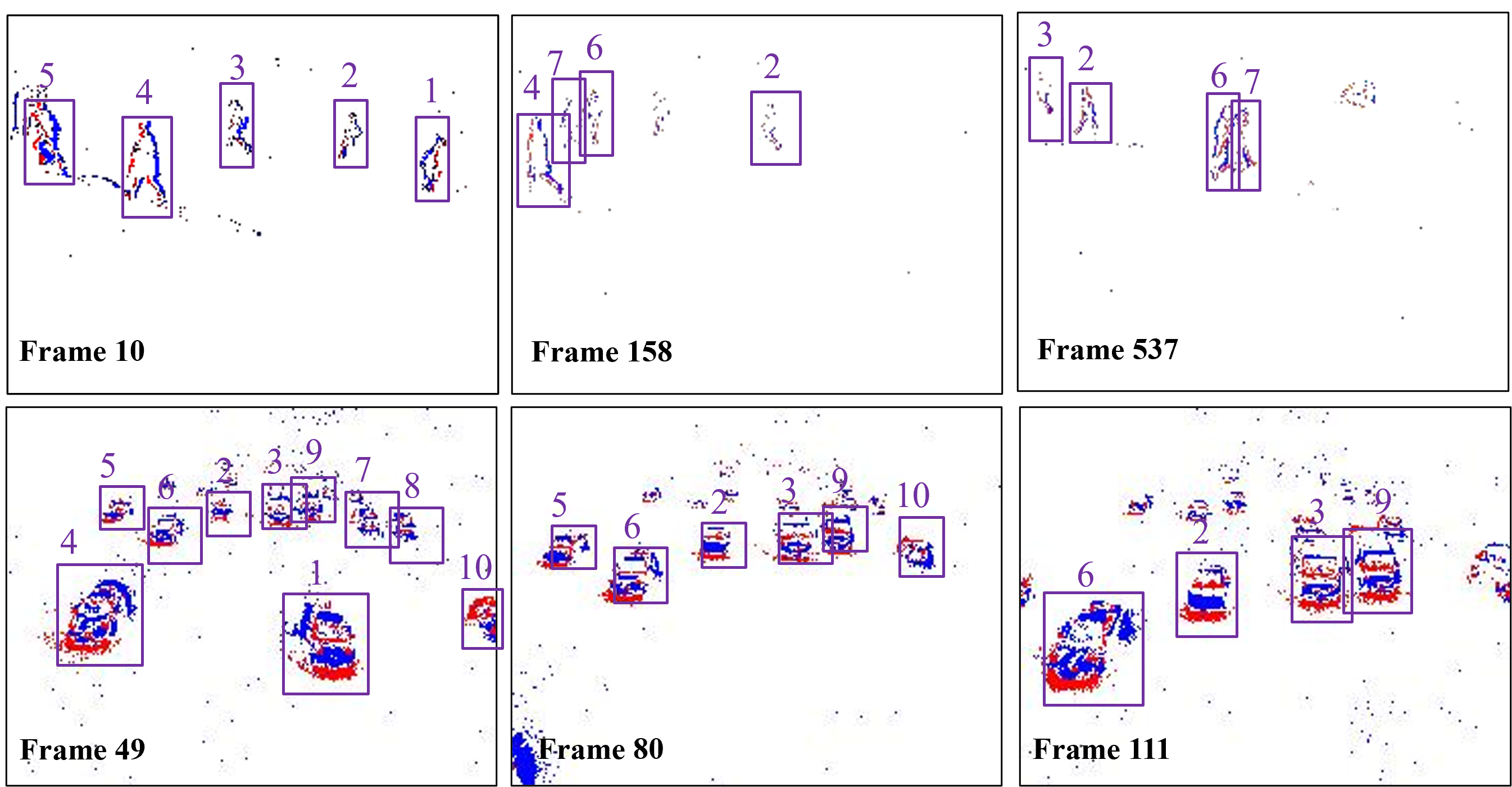}
    \caption{This visualization tracks the movement of pedestrians (10ms) and vehicles (20ms). In the first row, individuals with IDs 1, 2, 3, 4, and 5 are depicted walking from right to left, while those with IDs 6 and 7 move from left to right. In the second row, vehicles identified by IDs 7, 8, and 10 are shown driving from the foreground to the background, and others do the opposite direction.}
    \label{fig:tracking visualization}
\end{figure}

\fakeparagraph{Visualization analysis.}
We present the performance of our multi-object tracking benchmark. A selection of visualized results is shown in Fig.~\ref{fig:tracking visualization}. These illustrations depict pedestrians and vehicles, illustrating multi-object detection and tracking over time. Tables \ref{tab: Vehicle tracking performance} and \ref{tab: Pedestrian tracking performance} show the qualitative patterns align with the quantitative trends.

\subsection{Ablation studies}

\begin{table*}[t!]
\centering
\caption{Ablation study on the key component for Vehicle at different time intervals}
\begin{tabular}{@{\extracolsep{\fill}}c|cc|ccc|ccc|ccc|c|c@{}}
\toprule
\multicolumn{3}{c|}{Method} & \multicolumn{3}{c|}{EventSeq-Vehicle1} & \multicolumn{3}{c|}{EventSeq-Vehicle2} & \multicolumn{3}{c|}{EventSeq-Vehicle3} & \multicolumn{2}{c}{Speed}\\ 
\midrule
Time & MCD & Hybrid & MOTA & MOTP & IDF1 & MOTA & MOTP & IDF1 & MOTA & MOTP & IDF1 & mAP & FPS\\ 
\midrule
\multirow{3}{*}{10ms} &  &  & 26.7\% & \bfseries 28.8\% & 45.6\% & 25.0\% & \bfseries 29.5\% & 42.3\% & 13.9\% & \bfseries 31.7\% & 33.1\% & 36.9\% & 46 \\
 & \checkmark &  & 43.9\% & 21.9\% & 61.7\% & 56.2\% & 24.4\% & 60.1\% & 37.5\% & 25.4\% & 55.8\% & 39.1\% & 156\\
 & \checkmark & \checkmark & \bfseries 55.2\% & 21.4\% & \bfseries 74.5\% & \bfseries 56.2\% & 24.4\% & \bfseries 70.8\% & \bfseries 47.1\% & 25.6\% & \bfseries 64.4\% & 39.1\% & 198\\
\midrule
\multirow{3}{*}{20ms} &  &  & 5.1\% & \bfseries 22.6\% & 23.5\% & 21.6\% & 21.8\% & 36.8\% & 18.6\% & \bfseries 26.0\% & 33.5\% & 28.4\% & 47\\
 & \checkmark &  & 6.7\% & 22.5\% & 28.1\% & 26.8\% & 22.5\% & 42.9\% & 26.2\% & 24.7\% & 42.7\% & 26.6\% & 146\\
 & \checkmark & \checkmark & \bfseries 18.3\% & 23.5\% & \bfseries 51.9\% & \bfseries 58.4\% & \bfseries 25.2\% & \bfseries 73.2\% & \bfseries 51.7\% & 25.9\% & \bfseries 67.6\% & 26.6\% & 227\\
\midrule
\multirow{3}{*}{30ms} &  &  & 22.6\% & \bfseries 31.3\% & 41.9\% & 19.6\% & \bfseries 31.7\% & 39.2\% & 7.9\% & \bfseries 34.4\% & 28.6\% & 33.3\% & 97\\
 & \checkmark &  & 47.1\% & 25.6\% & 55.9\% & 52.5\% & 26.7\% & 59.3\% & 41.8\% & 28.0\% & 62.4\% & 30.5\% & 111\\
 & \checkmark & \checkmark & \bfseries 52.2\% & 21.6\% & \bfseries 65.8\% & \bfseries 58.8\% & 25.8\% & \bfseries 74.5\% & \bfseries 51.5\% & 28.6\% & \bfseries 70.5\% &30.5\% & 205\\
\bottomrule
\end{tabular}
\label{tab:ablation_veh}
\end{table*}

\begin{table*}[t!]
\centering
\caption{Ablation study on the key component for Pedestrians at different time intervals}
\begin{tabular}{@{\extracolsep{\fill}}c|cc|ccc|ccc|ccc|c|c@{}}
\toprule
\multicolumn{3}{c|}{Method} & \multicolumn{3}{c|}{EventSeq-Pedestrian1} & \multicolumn{3}{c|}{EventSeq-Pedestrian2} & \multicolumn{3}{c|}{EventSeq-Pedestrian3} & \multicolumn{2}{c}{Speed}\\ 
\midrule
Time & MCD & Hybrid & MOTA & MOTP & IDF1 & MOTA & MOTP & IDF1 & MOTA & MOTP & IDF1 & mAP & FPS\\ 
\midrule
\multirow{3}{*}{10ms} &  &  & 0.6\% & \bfseries 38.8\% & 5.3\% & \bfseries 7.7\% & 23.4\% & 12.6\% & 10.0\% & 26.9\% & 17.4\% & 15.3\% & 51\\
 & \checkmark &  & 0.4\% & 38.8\% & 4.3\% & 7.6\% & 25.2\% & 14.0\% & 4.3\% & 32.3\% & 17.7\% & 13.2\% & 132\\
 & \checkmark & \checkmark & \bfseries 17.5\% & 35.9\% & \bfseries 29.0\% & 7.5\% & \bfseries 32.1\% & \bfseries 31.0\% & \bfseries 16.3\% & \bfseries 33.2\% & \bfseries 29.4\% & 13.2\% & 210\\
\midrule
\multirow{3}{*}{20ms} &  &  & 5.1\% & 29.4\% & 9.9\% &8.6\% & 20.9\% & 16.0\% & 34.7\% & 26.0\% & 40.7\% & 25.0\% & 69\\
 & \checkmark &  & 14.7\% & 24.8\% & 28.1\% & 8.4\% & 30.4\% & 17.4\% & 38.0\% & 26.1\% & 44.7\% &25.6\% & 107\\
 & \checkmark & \checkmark & \bfseries 40.0\% & \bfseries 32.1\% & \bfseries 37.2\% & \bfseries 11.7\% & \bfseries 31.6\% & \bfseries 37.0\% & \bfseries 50.5\% & \bfseries 29.2\% & \bfseries 49.0\% & 25.6\% & 219\\
\midrule
\multirow{3}{*}{30ms} &  &  & 3.2\% & 30.2\% & 7.6\% & 9.0\% & 20.8\% & 16.8\% & 33.9\% & 26.7\% & 39.0\% & 23.8\% & 61\\
 & \checkmark &  & 26.7\% & \bfseries 32.7\% & 37.3\% & 15.4\% & \bfseries 31.6\% & 35.1\% & 35.7\% & \bfseries 30.4\% & 40.5\% & 22.5\% & 160 \\
 & \checkmark & \checkmark & \bfseries 40.0\% & 32.3\% & \bfseries 40.4\% & \bfseries 28.2\% & 29.4\% & \bfseries 44.3\% & \bfseries 45.4\% & 28.4\% & \bfseries 49.7\% & 22.5\% & 208\\
\bottomrule
\end{tabular}
\label{tab:ablation_ped}
\end{table*}

An ablation study assesses the impact of individual components within our tracking framework. Our ablation experiments are reported in Table~\ref{tab:ablation_veh} and Table~\ref{tab:ablation_ped}. By sequentially adding specific features and measuring the performance change, we quantify each component's contribution to the overall performance of our model. MCD denotes adding the MCD module to the detector, and Hybrid denotes a tracking strategy that uses both velocity change and direction.

Table~\ref{tab:ablation_veh} presents results on vehicles across three scenarios. The baseline model (no checkmarks) generally yields lower performance. Under the same experimental setup and evaluation protocol, incorporating MCD leads to a consistent runtime speedup while also improving accuracy. Meanwhile, adding Hybrid consistently increases tracking accuracy across all time windows. Overall, 20–30 ms favor tracking stability, while 10 ms favors detection for vehicles.

Table~\ref{tab:ablation_ped} reports the pedestrian ablation comparing tracking configurations across three scenarios (EventSeq-Pedestrian1/2/3) and three intervals (10ms, 20ms, 30ms). It demonstrates a similar trend. While MCD has limited impact on detection accuracy, Hybrid motion cues consistently boost pedestrian tracking performance. At 20--30ms, MOTA and IDF1 achieve their best levels, indicating that longer temporal windows combined with motion cues help maintain identity consistency in dense pedestrian scenes.

On MOTP, our final model does not always lead. The baseline sometimes reports artificially high MOTP, since tracking fewer objects reduces both \(d\) and \(M\). In contrast, our model attains higher IDF1, showing more reliable identity maintenance. This improvement is particularly evident at 20--30ms, where identity switches are fewer and long-term trajectories remain more coherent. The results highlight that our approach not only improves detection and tracking accuracy but also enhances robustness in preserving object identities. Overall, these findings confirm the complementary value of each component and the approach’s suitability for real-time vehicle and pedestrian tracking.

\section{Conclusion}

This paper presents TUMTraf EMOT, an event-based dataset for multi-object tracking (MOT) in Intelligent Transportation Systems (ITS). Event cameras, with high temporal resolution, dynamic range, and low latency, overcome the limitations of frame-based sensors in high-speed and low-light conditions. Our contributions include the TUMTraf EMOT dataset with diverse A9 traffic scenarios covering vehicles and pedestrians and a tracking-by-detection benchmark with a specialized feature extractor. The feature extractor is specifically designed for small-scale objects and crowded scenes in event-based imagery. Results demonstrate that event-based data by itself can reliably support multi-object tracking, even in challenging conditions. Future work will extend the dataset to broader scenarios and explore deep learning approaches for further gains in accuracy and adaptability.

\section*{Acknowledgement}
This work is supported by the MANNHEIM-CeCaS (Central Car Server – Supercomputing for Automotive, No. 16ME0820), in part by Tongji-Qomolo Autonomous Driving Commercial Vehicle Joint Lab Project, and in part by Xiaomi Young Talents Program.

\bibliographystyle{IEEEtran}
\bibliography{ref}


\end{document}